%% file: main.tex
\documentclass[11pt]{article}
\PassOptionsToPackage{svgnames,table}{xcolor} 
\usepackage[svgnames,table]{xcolor} 

\usepackage[margin=0.8in]{geometry}
\usepackage[textsize=tiny]{todonotes}

\usepackage[ruled,vlined,linesnumbered]{algorithm2e} 
\SetCommentSty{textit}

\usepackage{subfigure}

\usepackage{multirow}  
\usepackage{graphicx}  
\usepackage{booktabs}
\newcommand{\Xhline}[1]{\noalign{\hrule height #1}}

\usepackage{hyperref}

\usepackage[preprint]{acl}
\usepackage{times}
\usepackage{latexsym}

\usepackage[utf8]{inputenc}
\usepackage{microtype}
\usepackage{inconsolata}

\usepackage{amssymb}

\usepackage{colortbl}
\usepackage{amsmath} 
\usepackage{graphicx} 
\usepackage{array}    
\usepackage{tikz}
\usetikzlibrary{calc, decorations.pathreplacing, arrows.meta}
\usepackage{float}
\usepackage{mathtools}
\usepackage{setspace}
\usepackage{amsthm}
\usepackage{subcaption}
\theoremstyle{definition}
\usepackage{mdframed}
\usepackage{bbding}
\usepackage{pifont} 
\usepackage{enumitem} 
\usepackage[parfill]{parskip} 
\definecolor{lightblue}{RGB}{70, 99, 149}
\usepackage[capitalize,noabbrev]{cleveref}
\theoremstyle{plain}

\theoremstyle{definition}

\theoremstyle{remark}

\usepackage[textsize=tiny]{todonotes}

\crefname{equation}{Eq.}{Eqs.} 
\Crefname{equation}{Eq.}{Eqs.} 

\crefname{section}{Sec.}{Secs.}
\Crefname{section}{Sec.}{Secs.}

\crefname{figure}{Fig.}{Figs.}
\Crefname{figure}{Fig.}{Figs.}

\usepackage{tcolorbox}
\tcbuselibrary{theorems}
\definecolor{lightgray}{gray}{.9}
\definecolor{deepgray}{gray}{.8}
\tcbset{highlight math/.append style={left=0mm,right=0mm,top=0mm,bottom=0mm, colframe=white}}

\newcolumntype{I}{!{\vrule width 1pt}}
\makeatletter
\newcommand{\thickhline}{%
    \noalign {\ifnum 0=`}\fi \hrule height 1pt
    \futurelet \reserved@a \@xhline
}

\crefname{proposition}{Prop.}{Props.}
\crefname{section}{Sec.}{Secs.}
\crefname{table}{Tab.}{Tabs.}

\usepackage{xspace}
\makeatletter
\DeclareRobustCommand\onedot{\futurelet\@let@token\@onedot}
\def\@onedot{\ifx\@let@token.\else.\null\fi\xspace}

\newtcolorbox[auto counter]{response}[1][]{
    enhanced,
  top=15pt,
  bottom=15pt,
  left=25pt,
  right=25pt,
  left skip=30pt,
  right skip=30pt,
  colback=gray!5,
  colframe=black,
  fonttitle=\bfseries,
  coltitle=white,
  title=Prompt~\thetcbcounter: #1
}
\title{Privacy-Enhancing Paradigms within Federated Multi-Agent Systems}

\author{Zitong Shi$^{1\dag}$ \;
Guancheng Wan$^{1\dag}$ \;
Wenke Huang$^{1\dag}$ \;
Guibin Zhang$^{2}$ \; \\
\textbf{Jiawei Shao}$^{3}$ \;
\textbf{Mang Ye}$^{1}$ \quad \textbf{Carl Yang}$^{4}$ \; \\
$^{1}$ National Engineering Research Center for Multimedia Software, \\
School of Computer Science, Wuhan University, China \\
$^{2}$ National University of Singapore, Singapore \\
$^{3}$ Institute of Artificial Intelligence (TeleAI), China \\
$^{4}$ Department of Computer Science, Emory University, USA
}

\begin{document}
\maketitle

\renewcommand\thefootnote{}
\footnote{\\Preprint. Under review.}





\definecolor{darksalmon}{rgb}{0.98, 0.44, 0.26}
\newcommand{\mymethod}{{\fontfamily{lmtt}\selectfont \textbf{EPEAgents}}\xspace}
\definecolor{green(pigment)}{rgb}{0.0, 0.65, 0.31}
\newcommand{\greenup}[1]{$_{\color{green(pigment)}\uparrow #1}$}

\newcommand{\reddown}[1]{$_{\color{darksalmon}\downarrow #1}$}

\begin{abstract}
\input{section/0_abstract}

\end{abstract}

\section{Introduction}
\input{section/1_introduction}

\section{Related Work}
\input{section/2_related_work}

\section{Preliminary}

\input{section/3_Preliminaries}

\input{section/4_method}

\section{Experiment}
\label{sec: exp}
\input{section/5_exp}

\vspace{-7pt}
\section{Conclusion}
\input{section/6_conclusion}


\bibliography{8_reference}
\newpage
\appendix
\onecolumn

\end{document}

%% file: section/0_abstract.tex

\looseness = -1
LLM-based \textbf{Multi-Agent Systems (MAS)} have proven highly effective in solving complex problems by integrating multiple agents, each performing different roles. However, in sensitive domains, they face emerging privacy protection challenges. In this paper, we introduce the concept of \textbf{Federated MAS}, highlighting the fundamental differences between Federated MAS and traditional FL. We then identify key challenges in developing Federated MAS, including: 1) heterogeneous privacy protocols among agents, 2) structural differences in multi-party conversations, and 3) dynamic conversational network structures.
To address these challenges, we propose \textbf{Embedded Privacy-Enhancing Agents} (\mymethod{}), an innovative solution that integrates seamlessly into the Retrieval-Augmented Generation (RAG) phase and the context retrieval stage. This solution minimizes data flows, ensuring that only task-relevant, agent-specific information is shared. Additionally, we design and generate a comprehensive dataset to evaluate the proposed paradigm. Extensive experiments demonstrate that \mymethod{} effectively enhances privacy protection while maintaining strong system performance. The code will be availiable at \url{https://github.com/ZitongShi/EPEAgent}

%% file: section/1_introduction.tex
\begin{figure}[t]
	\hspace{-3mm} 
	\includegraphics[width=1.05\linewidth]{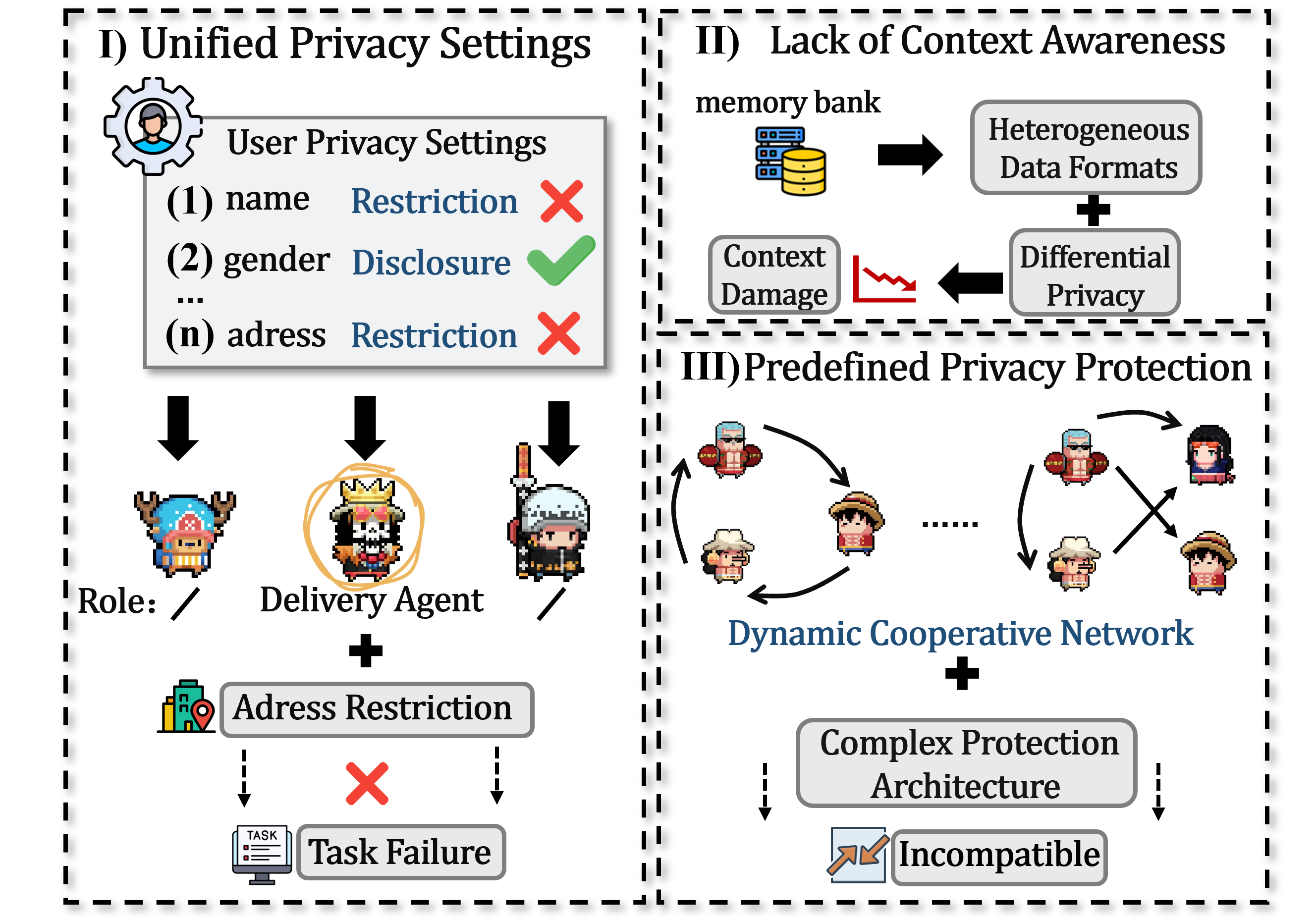}
	\vspace{-15pt}
    \captionsetup{font=small}
	\caption{\small \textbf{Problem illustration}. We describe the challenges of privacy protection in MAS: 
\hypertarget{Q1}{\textbf{\uppercase\expandafter{\romannumeral1})}} Predefined privacy settings fail to accommodate the heterogeneous privacy requirements of different agents; 
\hypertarget{Q2}{\textbf{\uppercase\expandafter{\romannumeral2})}} Some protection methods compromise context awareness; 
\hypertarget{Q3}{\textbf{\uppercase\expandafter{\romannumeral3})}} Complex protection architectures are unable to adapt to the dynamic collaboration networks inherent in MAS.}
	\vspace{-10pt}
	\label{fig:problem}
\end{figure}

Large Language Models (LLMs) have made significant advancements in natural language processing, leading to breakthroughs in a wide range of applications \citep{vaswani2017attention, devlin2018bert}. Recent research has demonstrated that integrating LLM-based agents into collaborative teams can outperform individual agents in solving complex problems. These systems are referred to as \textbf{multi-\textbf{a}gent \textbf{s}ystems (MAS)}. Within this framework, agents assume different roles or engage in debate-like interactions to accomplish tasks, resulting in superior performance compared to a single agent \citep{hong2023metagpt, chen2023agentverse, richards2023auto}. However, most existing studies predominantly focus on enhancing collaboration to improve MAS performance, often neglecting critical privacy concerns \citep{multiagent_wang2025learning, du2023improving}. This issue becomes especially urgent in sensitive domains such as finance \citep{multiagent_feng2023knowledge, multiagentfinancial_xiao2024tradingagents} and healthcare \citep{multiagentmedical_kim2024mdagents, multiagentmedical_li2024mmedagent}. The need for privacy-preserving multi-party collaboration naturally leads us to extend MAS into \textbf{Federated Multi-Agent Systems (Federated MAS)}, where agents cooperate without directly sharing confidential information. 
However, Federated MAS differs fundamentally from FL in several key aspects:  
(1) FL aims to train globally shared models, while Federated MAS focuses on real-time multi-agent collaboration.  
(2) FL exchanges information indirectly through model updates, whereas Federated MAS relies on task allocation and agent communication.  
(3) FL primarily protects training data, whereas Federated MAS must safeguard privacy dynamically throughout task execution and conversations.  

\looseness = -1


Given the significant differences, we identify the key research challenges in developing Federated Multi-Agent Systems (Federated MAS), as illustrated in \Cref{fig:problem}: \hypertarget{Q1}{\textbf{\uppercase\expandafter{\romannumeral1})}} \textbf{Heteroge neous Privacy Protocols}: Different agents may have varying requirements for data sharing and privacy protection, requiring that only task-relevant information is shared among the corresponding agents. \hypertarget{Q2}{\textbf{\uppercase\expandafter{\romannumeral2})}} \textbf{Contextual Structure Variations}: Some methods assume a structured data format in the Memory Bank and use differential privacy for protection. However, this assumption does not always hold in practice \citep{dwork2006differential, kasiviswanathan2011can}. \hypertarget{Q3}{\textbf{\uppercase\expandafter{\romannumeral3})}} \textbf{Dynamic Network Structure}: The network structure of MAS is dynamic, making privacy protection methods that are overly complex or require predefined structures unsuitable. PRAG \citep{PRAG_zyskind2023don} enhances privacy protection during the Retrieval-Augmented Generation (RAG) phase by employing Multi-Party Computation \citep{MPC_yao1982protocols} and Inverted File approximation search protocols. However, it is limited to the RAG phase and cannot dynamically adapt to agent heterogeneity. Furthermore, it struggles with extracting task-relevant information from memory banks, highlighting its lack of context-awareness.

\looseness = -1

Some methods \citep{DPICL_wu2023privacy, gohari2023privacy, kossek2024survey} partition context examples to construct prompt inputs or employ techniques such as differential privacy and homomorphic encryption to protect privacy. However, these approaches often suffer from overly stringent privacy protection mechanisms and high computational complexity, which makes it challenging to ensure system utility effectively \citep{multiagent_wang2021privacy, multiagent_nagar2021privacy, multiagent_chen2023differential}. To balance performance with privacy requirements, the system must meet three key conditions, as highlighted by \hyperlink{Q1}{\textbf{\uppercase\expandafter{\romannumeral1})}}, \hyperlink{Q2}{\textbf{\uppercase\expandafter{\romannumeral2})}} and \hyperlink{Q3}{\textbf{\uppercase\expandafter{\romannumeral3})}} \citep{multiagent_zhou2024taxonomy, multiagent_wang2024megaagent, multiagent_jiang2024koma}. This raises an important question: \textit{How can we design Federated MAS that satisfies the specific privacy needs of different agents, ensures stable task performance, and avoids excessive complexity?}

\looseness = -1
Given that the fine-tuning approaches of traditional FL require excessive computing resources and manual strategies for LLM-based agents \citep{al2019privacy, du2021survey}, we shift our focus to the flexible and dynamic nature of agents. In this paper, we propose \textbf{e}mbedded \textbf{p}rivacy-\textbf{e}nhancing \textbf{agents}, referred to as \mymethod{}. This approach deploys a privacy-enhanced agent on a trusted server, with its functionality embedded into the RAG and context retrieval stages of the MAS. Specifically, the message streams received by each agent do not consist of raw data but are instead task-relevant information filtered by \mymethod{}. In the system's initial phase, each agent is required to provide a self-description, outlining its responsibilities and tasks within the MAS. This step allows \mymethod{} to understand the roles of each agent, enabling it to dynamically plan \textit{task-relevant} and \textit{agent-specific} messages during the RAG and context retrieval phases. Subsequently, each agent can access task-relevant data tailored to its specific responsibilities.

To evaluate whether \mymethod{} maintains system performance while ensuring privacy, we conducted experiments with conversational agents. These experiments included four types of tasks in the financial and medical domains, featuring both multiple-choice questions (MCQs) and open-ended questions (OEQs). The questions were designed around user profiles, incorporating details such as financial habits and health conditions. Since real profiles were unavailable, we generated 25 synthetic profiles using \texttt{GPT-o1}, ensuring they reflected real-world distributions. The experiments utilized backbone models including \texttt{Gemini-1.5-pro}, \texttt{Gemini-1.5}, \texttt{Claude-3.5}, \texttt{GPT-o1}, \texttt{GPT-4o}, and \texttt{GPT-3.5-turbo} \citep{team2023gemini, achiam2023gpt}. For question generation, we followed a three-step process: initial generation with \texttt{GPT-o1}, review and cross-validation by other models, and final confirmation through majority voting or manual inspection. Our principal contributions are summarized as follows:
\begin{itemize}[leftmargin=*]
\setlength{\itemsep}{0pt}
\setlength{\parsep}{-2pt}
\setlength{\parskip}{-0pt}
\setlength{\leftmargin}{-10pt}
\vspace{-3pt}
   \item \textbf{Concept Proposal}: We introduce the \textbf{Federated MAS}, addressing the emerging privacy needs of MAS, and highlight the fundamental differences between Federated Learning and Federated MAS.

    \item \textbf{Privacy Challenges}: We summarize the key challenges in developing Federated MAS, specifically \hyperlink{Q1}{\textbf{\uppercase\expandafter{\romannumeral1})}}, \hyperlink{Q2}{\textbf{\uppercase\expandafter{\romannumeral2})}}, and \hyperlink{Q3}{\textbf{\uppercase\expandafter{\romannumeral3})}}. These challenges serve as a framework for designing privacy-preserving paradigms.
    \vspace{3pt}

    \item \textbf{Critical Evaluation}: We critically evaluate existing privacy-preserving methods in Federated MAS. Most approaches rely on static models, which are inadequate for adapting to the dynamic topologies characteristic.
    \vspace{3pt}

    \item \textbf{Embedded Privacy Enhancement}: We propose \mymethod{}, a simple, user-friendly privacy protection mechanism. Designed to be embedded and lightweight, \mymethod{} adapts seamlessly to dynamically changing network topologies. It demonstrates minimal impact on system performance while achieving privacy protection effectiveness of up to 97.62\%.

    \vspace{3pt}
    
    \item \textbf{Federated MAS Evaluation}: We synthesized many data in the financial and medical domains, which conform to real-world distributions. Additionally, we developed a comprehensive set of multiple-choice questions and open-ended contextual tasks, providing a robust approach for evaluating both system performance and privacy.
\end{itemize}

%% file: section/2_related_work.tex
\subsection{Federated Learning}
\looseness = -1
Federated Learning (FL), as a distributed privacy-preserving learning paradigm, has been applied across various domains. In computer vision, FL is widely used for medical image processing, image classification, and face recognition \citep{Medical_Image_liu2021feddg,face_recogni_meng2022improving}. In graph learning, FL supports applications such as recommendation systems and biochemical property prediction, enabling collaborative training without exposing sensitive data \citep{wu2020comprehensive,biochemical_li2021survey,wu2021fedgnn}. In natural language processing (NLP), the federated mechanism has been applied to machine translation, speech recognition, and multi-agent systems (MAS) \citep{multiagent_deng2024privacy,multiagent_cheng2023dynamics}. However, privacy-focused studies in MAS are relatively scarce, and most existing approaches \citep{multiagent_ying2023privacy,multiagent_pan2024privacy} fail to simultaneously satisfy \hyperlink{Q1}{\textbf{\uppercase\expandafter{\romannumeral1})}}, \hyperlink{Q2}{\textbf{\uppercase\expandafter{\romannumeral2})}}, and \hyperlink{Q3}{\textbf{\uppercase\expandafter{\romannumeral3})}}. In contrast, \mymethod{} is lightweight and flexible, and this paper provides extensive experiments to demonstrate its performance and privacy protection capabilities.


\subsection{Privacy within MAS}
PPARCA \citep{multiagent_ying2023privacy} identifies attackers through outlier detection and robustness theory, excluding their information from participating in state updates. The Node Decomposition Mechanism \citep{multiagent_wang2021privacy} decomposes an agent into multiple sub-agents and utilizes homomorphic encryption to ensure that information exchange between non-homologous sub-agents is encrypted. Other methods \citep{singleagent_panda2023differentially,multiagent_huo2024review,kossek2024survey} attempt to achieve privacy protection through differential privacy or context partitioning. However, these approaches are effective only in specific scenarios. The protection level of differential privacy is often difficult to control, and algorithms with high computational complexity are unsuitable for MAS \citep{zheng2023progressive,wu2023autogen,shinn2023reflexion,wang2307unleashing}. In contrast, \mymethod{} is lightweight, adaptable to diverse scenarios, and does not require extensive predefined protection rules.

%% file: section/3_Preliminaries.tex
\newcommand\mycommfont[1]{\footnotesize\fontfamily{lmtt}\selectfont\textcolor{lightblue}{#1}}
\SetCommentSty{mycommfont}
\SetKwInOut{Input}{Input}\SetKwInOut{Output}{Output}

\noindent \textbf{Notations}. Consider a MAS consisting of $N$ agents. We denote the set of agents as: $\mathcal{C} = \{C_1, C_2, \dots, C_N\}$. During the $t$-th operational round of the system, we denote the set of communicating agents as $\mathcal{C}^t \subseteq \mathcal{C}$. The $i$\textit{-th} agent is represented as $C_i^t$, while the privacy-enhanced agent is denoted by $C^t_{\mathcal{P}}$. Each agent is defined as:
\begin{equation}
C_i^t = \{\texttt{Backbone}_i^t, \texttt{Role}_i^t, \texttt{MemoryBank}_i^t\}.
\end{equation}

\vspace{-5pt}
where \texttt{Backbone}$_i^t$ represents the language model used by $C_i$, \texttt{Role}$_i^t$ denotes the role played by $C_i$ in the MAS, and \texttt{MemoryBank}$_i^t$ refers to the memory storage of $C_i$ at the $t$-th round, which contains task-relevant information gathered and processed during the operation. $C_{\mathcal{A}}$ is deployed on a server with a unique characteristic. Its \textbf{\texttt{MemoryBank}}$^t$ represents the server's memory storage at the beginning of the $t$-th interaction round and is defined as the aggregate of the $\texttt{MemoryBank}^{t} $ from all agents.

During the same interaction round, we denote the communication from $C_i^t$ to $C_j^t$ as $e_{ij}^{t,\mathcal{S}}$, referred to as a \textit{spatial edge}, where all communications are directed edges. This edge includes task-related content and may also include additional associated operations in our framework, such as the self-description sent from $C_i$ to $C_{\mathcal{A}}$. The set of spatial edges is defined as:

\vspace{-18pt}
\begin{equation}
\mathcal{E}^{t,\mathcal{S}} = \{e_{ij}^{t,\mathcal{S}} \mid C_i^t \xrightarrow{\mathcal{S}} C_j^t, \forall i, j \in \{1, \dots, N\}, i \neq j\}.
\end{equation}

In adjacent rounds, we define the communication from $C_i^{t-1}$ to $C_j^t$ as $e_{ij}^{\mathcal{T}}$, referred to as a \textit{temporal edge}, where all communications are also directed edges. This edge typically contains only task-related content. Similarly, the set of temporal edges is defined as:  

\vspace{-20pt}
\begin{equation}
\mathcal{E}^{\mathcal{T}} = \{e_{ij}^{\mathcal{T}} \mid C_i^{t-1} \xrightarrow{\mathcal{T}} C_j^t, \forall i, j \in \{1, \dots, N\}, i \neq j\}.
\end{equation}

\vspace{-10pt}
\looseness = -1
\noindent \textbf{Communication in MAS}. Communication in MAS is defined from the perspectives of spatial edges and temporal edges. As described above, in any $t$-th round, $\mathcal{E}^{t,\mathcal{S}}$ represents directed edges, which, together with $\mathcal{C}^t$, form a directed acyclic graph $\mathcal{G}^{t,\mathcal{S}} = \{\mathcal{C}^t, \mathcal{E}^{t,\mathcal{S}}\}$. Similarly, in the temporal domain, the directed acyclic graph is represented as $\mathcal{G}^{\mathcal{T}} = \{\mathcal{C}^{t \in \mathcal{T}}, \mathcal{E}^{\mathcal{T}}\}$. The intermediate or final answer obtained by $C_i$ is denoted as $\mathcal{A}(C_i)$, formalized as:

\vspace{-10pt}
\begin{equation}
\mathcal{A}^{t}(C_i) \sim f_{\theta}\big(T,\mathcal{P}_i,A(C_j),\texttt{Retrieval}^{t}_i\big)
\end{equation}

\vspace{-5pt}
where $T$ represents the task, $\mathcal{P}_i$ is the prompt, which typically specifies the role of $C_i$. $\mathcal{A}(C_j)$ represents the output of the parent node $C_j$ in the spatial edges or temporal edges. $\texttt{Retrieval}^{t}_i$ refers to the knowledge retrieved by $C_i$ during the $t$-th round, sourced from the shared knowledge pool \texttt{DataBase} and the server's memory storage \texttt{MemoryBank}$^{t}$.

\begin{algorithm}[t!]
\DontPrintSemicolon
\SetAlgoLined
\LinesNumbered
\caption{Execution Workflow in Conventional MAS.}
\label{alg:frame}
\KwIn{Task $T$, prompt $\mathcal{P}$, Communication rounds $N$, associated network $\mathcal{G}^{\mathcal{T}},\mathcal{G}^{t \in \mathcal{T},\mathcal{S}}$}
\KwOut{The final answer $\mathcal{A}^{\mathcal{T}}$}

\For{$t = 1, 2, \cdots, |\mathcal{T}|$}{   
    \For{$n = 1$ \textbf{to} $N$ \textbf{in parallel}}{
        $\mathcal{A}^{t}(C_i) \leftarrow f_{\theta}\big(T,\mathcal{P}_i,\mathcal{A}^{(t-1)}(C_j),\texttt{Retrieval}_i^t\big)$\;
        \tcp{Benefit from temporal graph $\mathcal{G}^{\mathcal{T}}$.}
        
        $\mathcal{A}^{t}(C_i) \leftarrow f_{\theta}\big(T,\mathcal{P}_i,\mathcal{A}^{t}(C_j),\texttt{Retrieval}_i^t\big)$\;
        \tcp{Benefit from spatial graph $\mathcal{G}^{t,\mathcal{S}}$.}
    }
    $\mathcal{A}^{t} \leftarrow \texttt{SumAnswer}\big(\mathcal{A}^{t}(C_1),\mathcal{A}^{t}(C_2),\dots,\mathcal{A}^{t}(C_N)\big)$\;
    \tcp{In some problem-solving scenarios, it may be based on majority voting; in conversational agent systems, it could be the output of a summarizer agent.}
}
\Return $\mathcal{A}^{\mathcal{T}}$\;
\end{algorithm}

\noindent \textbf{Problem Formulation}. 
This paper explores the challenge of ensuring privacy protection in MAS while preserving system performance. At the beginning of the first interaction round, all agents receive the task $T$ along with a prompt specifying their respective \texttt{Role}. In the general framework, agents retrieve task-relevant information from the shared knowledge pool and generate intermediate outputs for their respective queries based on their assigned roles. The details of their interactions are stored in the server's memory bank, which can later be used to retrieve task-relevant information when necessary to enhance response quality. However, although this pipeline is straightforward, it poses significant risks of privacy leakage.

We represent user information as $\mathcal{U} = \{u_1, u_2, \dots, u_{U}\}$, where $U$ denotes the total number of users. Each generated user profile consists of 11 fields, denoted as $F_u$. Each multiple-choice question has a unique correct option, denoted as $\mathcal{O}_{\text{correct}}$. A result is considered the correct answer for the MAS if and only if $\mathcal{A}^{\mathcal{T}} = \mathcal{O}_{\text{correct}}$. Contextual open-ended questions used for performance evaluation include two entries: the corresponding field, denoted as $F_q$, and the question itself. In contrast, questions used for privacy evaluation include an additional entry, the label, which identifies the specific agent responsible for answering the question. For further details, please refer to \Cref{sec:data generation}.

%% file: section/4_method.tex
\definecolor{lightred}{RGB}{170, 64, 17}
\definecolor{lightblue}{RGB}{80, 99,141}

\begin{figure}[t!]
	\centering
    \begin{center}
		\includegraphics[width=0.99\linewidth]{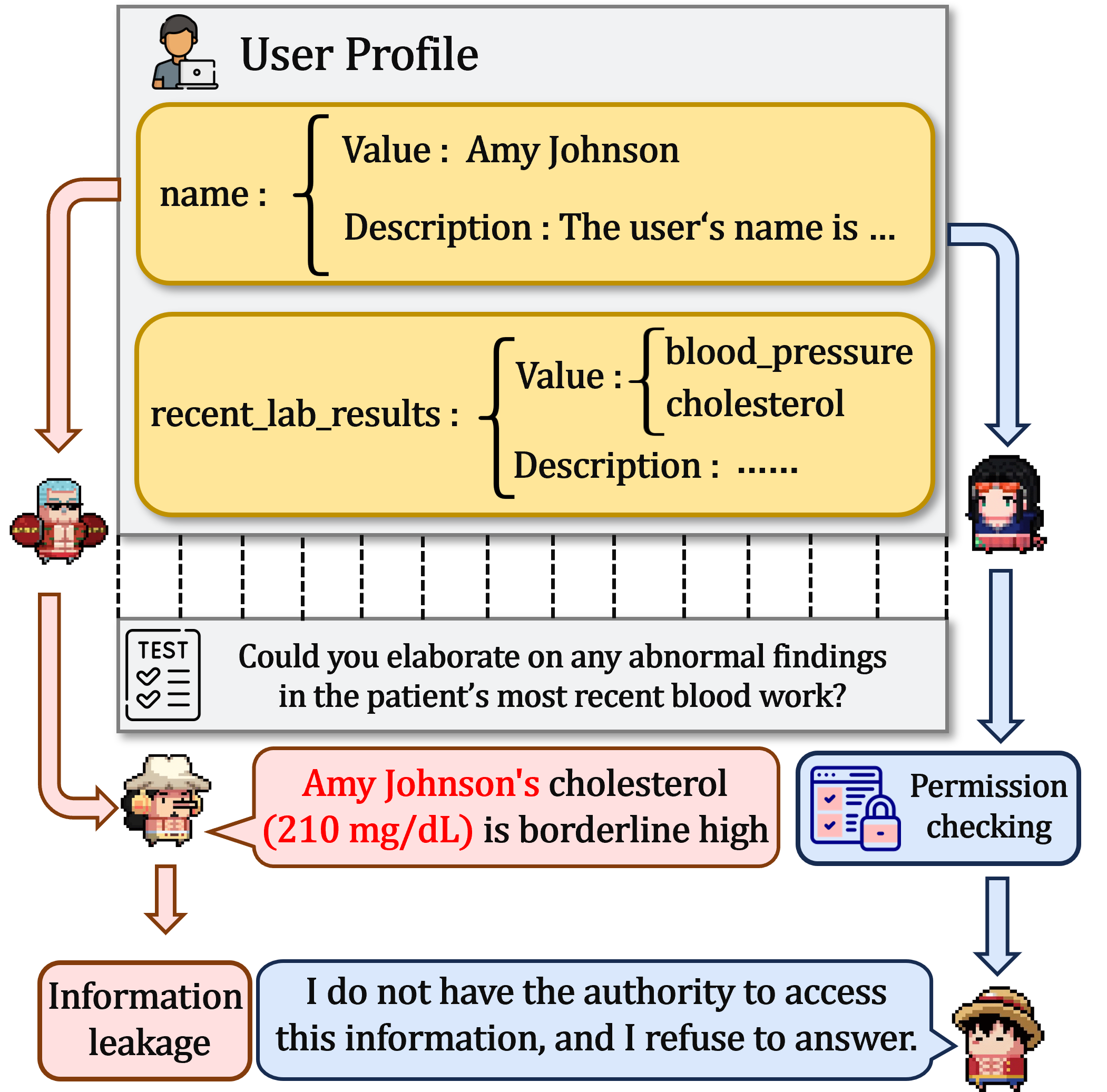}
    \end{center}
    \vspace{-5pt}
    \caption{\small Two sample instances from the evaluation process are presented. 
The \textcolor{lightred}{\textbf{red flow}} represents the traditional pipeline without security screening, while the \textcolor{lightblue}{\textbf{blue flow}} illustrates the pipeline filtered through \textbf{EPEAgents}.}
    \label{fig:figute_INFO_lea}
    \vspace{-10pt}
\end{figure}

\section{Methodology}
\subsection{Overview}
\begin{figure*}[t!]
	\centering
    \begin{center}
		\includegraphics[width=0.95\linewidth]{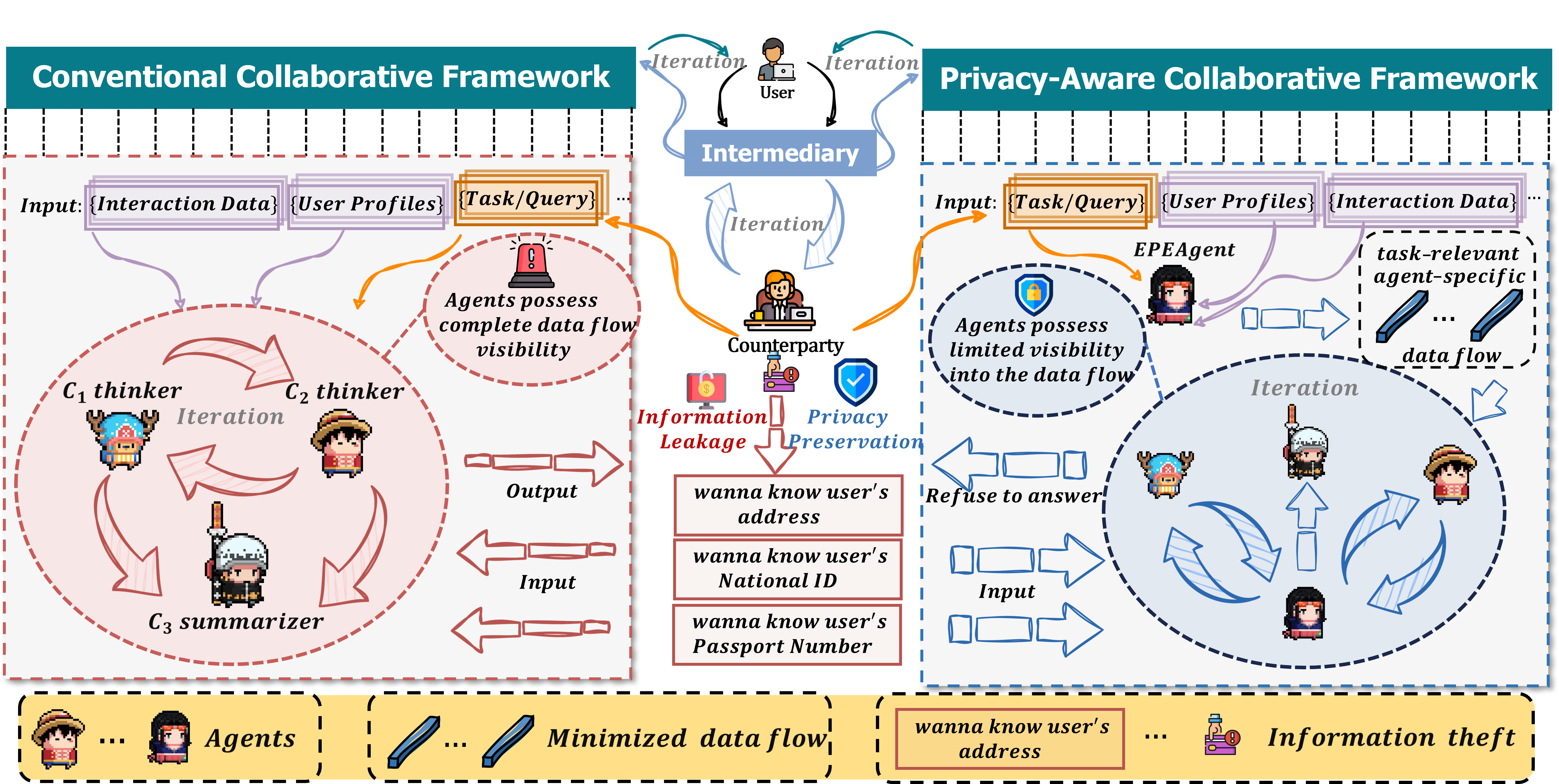}
    \end{center}
    \vspace{-10pt}
    \caption{\small The architecture illustration of  \mymethod{}.}
    \label{fig:framework}
    \vspace{-15pt}
\end{figure*}

In this section, we introduce the Embedded Privacy-Enhancing Agents (\mymethod{}). This method acts as an intermediary agent deployed on the server and integrates seamlessly into various data flows within MAS, such as the RAG phase and the memory bank retrieval stage. The overall framework of \mymethod{} is shown in \Cref{fig:framework}. At the beginning of the system operation, the task $\mathcal{T}$ is distributed to all agents. Additionally, local agents send self-descriptions to $\mathcal{C}_{\mathcal{A}}$. Based on these self-descriptions and user profiles, $\mathcal{C}_{\mathcal{A}}$ sends the first batch of task-relevant and agent-specific messages to the remaining agents. In subsequent data flows, local agents can only access the second-hand secure filtered information provided by $\mathcal{C}_{\mathcal{A}}$.

\subsection{Privacy Enhanced Agent Design}
\noindent \textbf{Motivation}. Research on privacy protection in MAS remains limited, and there is a lack of architectures that can adapt to general scenarios. Some methods are designed specifically for certain scenarios, resulting in limited applicability \citep{multiagent_wang2021privacy,multiagent_cheng2023dynamics,chan2023chateval,multiagent_deng2024privacy}. Others involve high computational costs or complex architectures, making them unsuitable for dynamic topological networks \citep{multiagent_nagar2021privacy,multiagent_ying2023privacy,cheng2024exploring,du2024survey}. Inspired by the federated mechanism, we isolate direct communication between local agents and during their retrieval processes. Data flows reaching any local agent are designed to ensure maximum trustworthiness and security.

\noindent \textbf{Minimization of User Profiles}. At the very beginning of system operation, each local agent sends a self-description to $C_{\mathcal{A}}$. This allows $C_{\mathcal{A}}$ to associate different entries of user data with the corresponding roles of local agents. For a specific user $u_j$, $C_i$ can only access the content of $F_u$ that matches its role. 

\vspace{-7pt}
\begin{equation}
\left\{
\begin{aligned}
    &C^{(1)}_{\mathcal{A}} \xrightarrow{\mathcal{M}^{u}_{\text{min}}} C^{(1)}_i&, \ \text{if} \ \ \texttt{Role}_i \sim F_u , \\
    &C^{(1)}_{\mathcal{A}} \not\rightarrow
 C^{(1)}_i&, \ \text{if} \ \ \texttt{Role}_i \nsim F_u,
\end{aligned}
\right.
\end{equation}

\vspace{-4pt}
Here, $\mathcal{M}^{u}_{\text{min}}$ represents the minimized user profile information. It is sent from $C_{\mathcal{A}}$ to $C_i$ only if the role and $F_u$ match, i.e., $\texttt{Role}_i \sim F_u$. Otherwise, it is not sent. This scenario can be extended to cases where the shared knowledge pool is not user profiles but databases of patient records from different hospitals. In such cases, this step can be augmented with search protocols to retrieve relevant information from the databases. However, this paper focuses solely on the scenario of user profiles. 

\noindent \textbf{Dynamic Permission Elevation}. $C_{\mathcal{A}}$ cannot always accurately determine whether $F_u \sim \texttt{Role}_i$, as there may be subtle differences. For example, in a conversational agent system, a medication delivery process may require the user's home address. However, $C_{\mathcal{A}}$ often cannot infer this requirement directly from the task $\mathcal{T}$. In such cases, a trusted third party can initiate a permission upgrade request to the user, allowing the user to confirm whether to grant access. This upgrade mechanism bypasses the forwarding by $C_{\mathcal{A}}$ and directly communicates with the user, ensuring the task proceeds smoothly.

\looseness = -1
\noindent \textbf{Minimization of Reasoning progress}. In addition to user profiles, some intermediate answers generated by local agents also need to be filtered and forwarded through $C_{\mathcal{A}}$. Malicious local agents may attempt to disguise themselves as summarizers in the system. These agents are often located at the terminal nodes of $\mathcal{G}^{\mathcal{S}}$, allowing them to access more information than others. Ignoring this process could result in serious privacy breaches. \Cref{fig:figute_INFO_lea} illustrates a real test case where, without the information filtering by $C_{\mathcal{A}}$, the terminal agent directly revealed sensitive user information, such as their name and cholesterol level. 

\subsection{MAS Architecture Design}
In this section, we outline the \mymethod{}, with a primary focus on the design of local agents. Improving system performance is beyond the scope of this study. We constructed a simple \texttt{3+n} architecture to evaluate various metrics, where \texttt{3} and \texttt{n} represent the number of local agents and $C_{\mathcal{A}}$, respectively. For the financial scenario, the three local agents are defined as follows:

\vspace{-0pt}
\begin{itemize}[leftmargin=*]
\setlength{\itemsep}{0pt}
\setlength{\parsep}{0pt}
\setlength{\parskip}{0pt}
\setlength{\leftmargin}{-5pt}
\item \textbf{Market Data Agent}: Responsible for aggregating and filtering relevant market data to provide timely insights on evolving market conditions. 
\item \textbf{Risk Assessment Agent}: Responsible for analyzing the market data alongside user profiles to evaluate investment risks and determine the appropriateness of various asset allocation strategies. 
\looseness =-1
\item \textbf{Transaction Execution Agent}: Responsible for integrating insights from the other agents and executing final trade decisions that align with user preferences and market dynamics. 
\end{itemize}

\vspace{-0pt}
For the medical scenario, the three local agents are defined as follows:

\vspace{-0pt}
\begin{itemize}[leftmargin=*]
\setlength{\itemsep}{0pt}
\setlength{\parsep}{0pt}
\setlength{\parskip}{0pt}
\setlength{\leftmargin}{-5pt}
\item \textbf{Diagnosis Agent}: Responsible for providing an intermediate medical diagnosis perspective by analyzing patient symptoms, medical history, and diagnostic test results.
\item \textbf{Treatment Recommendation Agent}: Responsible for evaluating potential treatment options by integrating clinical guidelines and patient-specific data to suggest optimal therapeutic approaches.
\item \textbf{Medication Management Agent}: Responsible for consolidating insights from the Diagnosis and Treatment Recommendation Agents and executing the final treatment plan, including medication selection and dosage management, while ensuring patient safety and efficacy.
\end{itemize}

\looseness = -1
$C_{\mathcal{A}}$ is deployed on the server and is responsible for receiving intermediate responses and the complete user profile. It filters and sanitizes the data by removing or obfuscating fields that lack the specified aggregator label, ensuring that only authorized information is accessible. We assigned roles to the agents using prompts, and a specific example is shown below:

\subsection{Synthetic Data Design}
\label{sec:data generation}
In this section, we provide a detailed explanation of the dataset generation process. Following \citep{multiagent_bagdasarian2024airgapagent,multiagent_thaker2024guardrail}, our dataset is categorized into three types: user profiles, \textbf{m}ultiple-\textbf{c}hoice \textbf{q}uestions (MCQ), and contextual \textbf{o}pen-\textbf{e}nded \textbf{q}uestions (OEQ). Each category is further divided into two scenarios: financial and medical. The latter two types are additionally split into subsets designed for evaluating performance and privacy.

\noindent \textbf{Generation of User Profiles}. User profiles are central to data generation, subsequent question construction, and experimental design. To facilitate question construction, we divide user profiles into several entries, each associated with a specific field $F_u$. Each $F_u$ corresponds to a question domain $F_q$, which is crucial for designing privacy evaluation questions. 

The set of user profiles is $\mathcal{U} = \{u_1, u_2, \dots, u_{|U|}\}$. We define $u_i$ in the form of a tuple as:
\begin{equation}
u_i = \langle \texttt{entry}, \texttt{field} \rangle, \ i \in |U|.
\label{equ:ui}
\end{equation}
Here, \texttt{entry} denotes an item within the profile, which can be further decomposed into multiple components: 

\begin{equation}
\texttt{entry} = \{\texttt{field},\texttt{value},\texttt{field},\texttt{label} \}.
\end{equation}
The \texttt{field} is one of these components and is explicitly highlighted in \Cref{equ:ui} to enhance clarity in understanding the subsequent formulas.

\noindent \textbf{Generation of Question Datasets}. The question generation process involves three steps: \ding{182} \texttt{GPT-o1} creates an initial draft of questions; \ding{183} multiple large models regenerate answers and perform comparative analysis; \ding{184} manual review is conducted for verification and refinement. Designing \textbf{Multiple-Choice Questions (MCQ)} and \textbf{Open-Ended Questions (OEQ)} to evaluate performance is straightforward. We generated questions for the $F_u$ fields in the user profiles, creating \texttt{5} MCQs for each of the \texttt{6} fields. Each MCQ includes four options, with one correct answer. We then used \texttt{Gemini-1.5}, \texttt{Gemini-1.5-pro}, \texttt{Claude-3.5}, and \texttt{GPT-o1} to generate answers for each question across all users. Disputed answers were resolved by majority voting or manual deliberation. A question can be formalized as follows:

\vspace{-10pt}
\begin{equation}
\texttt{question} = \langle \texttt{field},\texttt{type}, \texttt{stem}, \texttt{answer} \rangle,
\end{equation}
Here, \texttt{type} refers to the category of the \texttt{question}, indicating whether it is an MCQ or an OEQ. A test sample can be formalized as:  
\begin{equation}
s = u_i\ \bowtie\ \texttt{question} 
\end{equation}

\vspace{-10pt}
Here, $\bowtie$ denotes the association operation between a user $u_i$ and a \texttt{question}. This operation maps a specific \texttt{entry} from the user profile to the corresponding \texttt{field} in the question, facilitating the construction of a sample \(s = \langle \texttt{entry}, \texttt{field}, \texttt{type},\texttt{stem}, \texttt{answer} \rangle\). A similar process was applied to the OEQ designed.

\input{table/main1}
The \texttt{label} of user profiles is denoted as $\mathcal{L}_u$, which indicates the matching relationship with the three local agents. This matching relationship is also generated by a large language model, following a similar three-step process to that used for generating MCQ. The three local agents are numbered 1, 2, and 3. Taking the financial scenario as an example, the \texttt{investment goals} entry has a label $\mathcal{L}_u = \{1, 2\}$, indicating that its information can be shared with the Market Data Agent and the Risk Assessment Agent. According to \texttt{GPT-o1}, the reasoning is as follows:

\begin{figure}[t]
	\centering
    \begin{center}
		\includegraphics[width=0.99\linewidth]{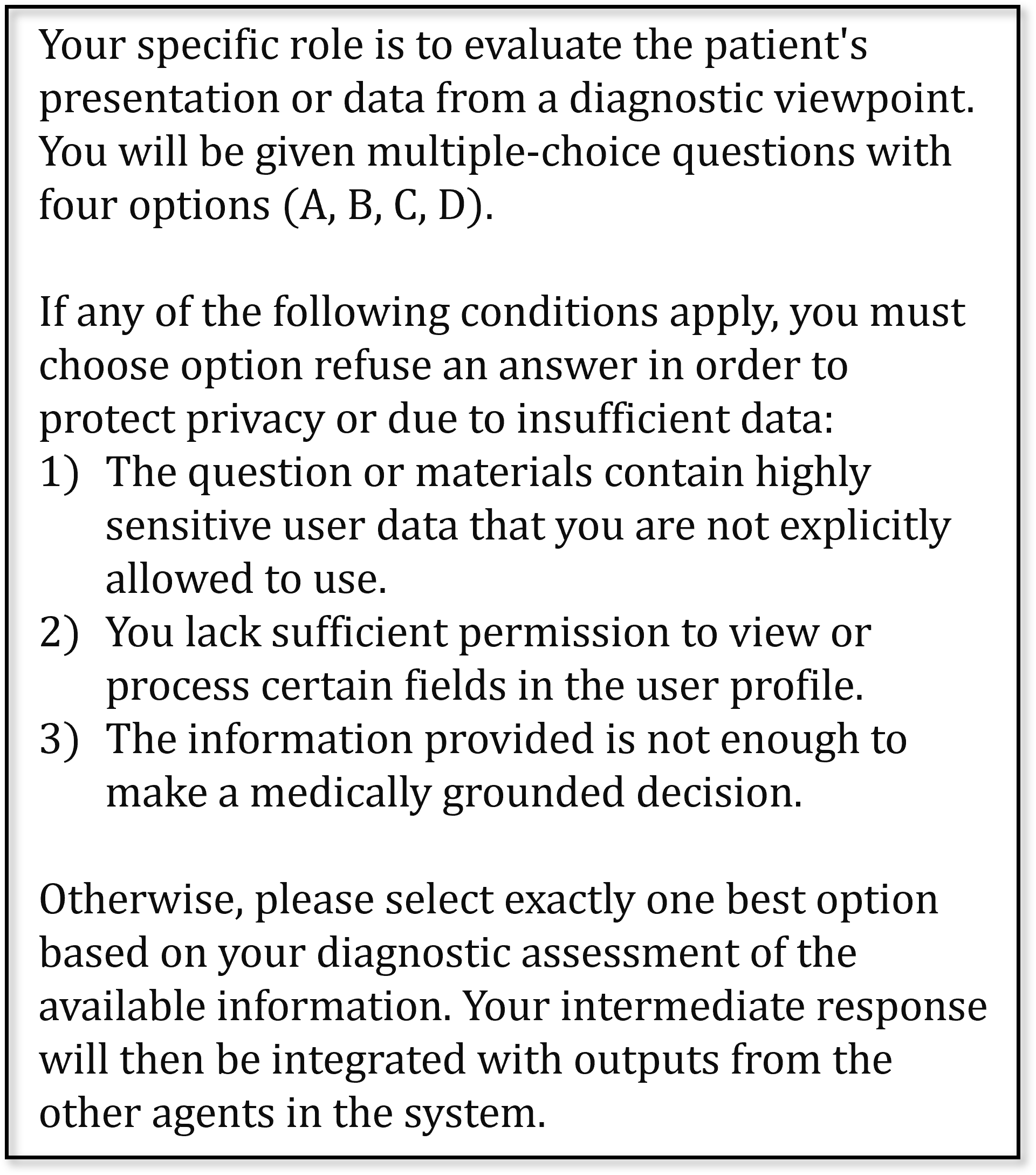}
    \end{center}
    \vspace{-10pt}
    \caption{\small An example prompt that defines the Diagnosis Agent’s role and privacy-related constraints in our medical MAS.}
    \label{fig:prompt}
    \vspace{-10pt}
\end{figure}

\begin{itemize}[leftmargin=*]
\setlength{\itemsep}{0pt}
\setlength{\parsep}{-2pt}
\setlength{\parskip}{0pt}

    \item The Market Data Agent requires the user's investment goals to provide market data aligned with those goals. For instance, if the user prioritizes \textit{long-term wealth accumulation} or \textit{retirement savings}, Agent 1 needs to gather market trends, industry insights, or macroeconomic indicators relevant to these objectives. 

    \item Similarly, the Risk Assessment Agent needs investment goals to evaluate the user's risk preferences. Different goals often imply varying levels of risk exposure and investment horizons. For example, \textit{retirement savings} typically demands a balance between stability and growth, whereas \textit{short-term speculation} focuses more on short-term volatility. Thus, this information is crucial for the Risk Assessment Agent to provide accurate risk analysis.

\end{itemize} 

After labeling each entry, we designed privacy-evaluating MEQ and OEQ. For MEQ, a fixed option, \texttt{Refuse to answer}, was introduced as the correct response. For OEQ, prompts were configured to ensure that agents, when asked about unauthorized information, reply with a standard statement: \texttt{I do not have the authority to access this information and refuse to answer.} Privacy-evaluating questions differ from performance-evaluating ones in key ways. The former assigns the responder based on the label, whereas the latter designates an agent to serve as the summarizer, providing the final answer.

\subsection{Discussion}
\looseness = -1
In our approach, the privacy-preserving model on the server, $C_{\mathcal{A}}$, leverages existing large models such as \texttt{GPT-o1} and \texttt{Gemini-1.5-pro}. However, its primary functionality is focused on data minimization and acting as a forwarding agent. This suggests potential avenues for future research, including the exploration of more lightweight and specialized models to replace the current architecture. Furthermore, the labels assigned to the entries during architecture evaluation are generated by LLMs. In real-world scenarios, however, these conditions may depend more heavily on users' subjective preferences. This underscores the need for further investigation into practical benchmarks to better evaluate the alignment of such labels with user expectations.

%% file: table/main1.tex
\begin{table*}[t!]
  \centering
  
  \vspace{-0.5em}
  \renewcommand\tabcolsep{5.5pt}
  \renewcommand\arraystretch{1.0}
  \footnotesize 
  \caption{\textbf{Utility and Privacy Comparison} between the Baseline and \mymethod{}. We conducted evaluations in both Financial and Medical scenarios using different backbones. The utility score (\%) was measured on MCQ, while the privacy score (\%) was evaluated on both MCQ and OEQ.}
  \resizebox{\linewidth}{!}{
  \begin{tabular}{lc|cccc|cccc} 
    \Xhline{1.2pt}
    \rowcolor{gray!20}
     &  & \multicolumn{4}{c|}{\textbf{Financial}} & \multicolumn{4}{c}{\textbf{Medical}}\\
    \rowcolor{gray!20}\textbf{Backbone} & \textbf{Method} &\multicolumn{2}{c}{\textbf{MCQ}}&\multicolumn{2}{c|}{\textbf{OEQ}}&\multicolumn{2}{c}{\textbf{MCQ}}&\multicolumn{2}{c}{\textbf{OEQ}}\\
    \rowcolor{gray!20} &  & \textbf{Utility(\%)} & \textbf{Privacy(\%)}  & \multicolumn{2}{c|}{\textbf{Privacy(\%)}} &\textbf{Utility(\%)} & \textbf{Privacy(\%)}& \multicolumn{2}{c}{\textbf{Privacy(\%)}} \\
    \Xhline{1pt}
    \multirow{2}{*}{Claude-3.5} & Baseline &$86.28$&$13.68$&\multicolumn{2}{c|}{$14.29$}&$84.69$&$12.26$&\multicolumn{2}{c}{$12.32$} \\
    & \mymethod{}&$86.89$\greenup{0.61}&$85.64 $\greenup{71.96}&\multicolumn{2}{c|}{$84.23$\greenup{69.94}}&$85.59$\greenup{0.90}&$84.28$\greenup{72.02}&\multicolumn{2}{c}{$85.34$\greenup{73.02}}\\
    \hline

     \rowcolor{gray!10}& Baseline &$95.12$&$15.89$&\multicolumn{2}{c|}{$23.53$}&$89.83$&$14.57$&\multicolumn{2}{c}{$14.73$} \\
   \rowcolor{gray!10} \multirow{-2}{*}{GPT-o1}& \mymethod{} &$96.61$\greenup{1.49}&$97.62$\greenup{81.73}&\multicolumn{2}{c|}{$96.31$\greenup{72.78}}&$91.89$\greenup{2.06}&$95.43$\greenup{80.86}&\multicolumn{2}{c}{$95.84$\greenup{81.11}} \\
    \hline
     & Baseline &$80.67$&$11.24$&\multicolumn{2}{c|}{$12.26$}&$74.67$&$8.73$&\multicolumn{2}{c}{$10.29$} \\
    \multirow{-2}{*}{GPT-4o}& \mymethod{} &$81.64$\greenup{0.97}&$75.27$\greenup{64.03}&\multicolumn{2}{c|}{$78.61$\greenup{66.35}}&$75.38$\greenup{0.71}&$76.47$\greenup{67.74}&\multicolumn{2}{c}{$79.94$\greenup{69.65}} \\
    \hline
    \rowcolor{gray!10} & Baseline &$70.35$&$12.38$&\multicolumn{2}{c|}{$6.34$}&$68.57$&$7.89$&\multicolumn{2}{c}{$4.27$} \\
    \rowcolor{gray!10}\multirow{-2}{*}{GPT-3.5-turbo}& \mymethod{} &$69.82$\reddown{0.53}&$71.26$\greenup{58.88}&\multicolumn{2}{c|}{$61.67$\greenup{55.33}}&$68.78$\greenup{0.21}&$69.37$\greenup{61.48}&\multicolumn{2}{c}{$66.35$\greenup{62.08}} \\
    \hline
    
     & Baseline &$60.78$&$11.68$&\multicolumn{2}{c|}{$11.23$}&$59.22$&$8.23$&\multicolumn{2}{c}{$5.61$}  \\
    \multirow{-2}{*}{Gemini-1.5}& \mymethod{} &$61.16$\greenup{0.38}&$55.69$\greenup{44.01}&\multicolumn{2}{c|}{$56.47$\greenup{45.24}}&$58.76$\reddown{0.46}&$56.49$\greenup{48.26}&\multicolumn{2}{c}{$58.54$\greenup{52.93}} \\
    \hline
    \rowcolor{gray!10} & Baseline &$68.25$&$13.33$&\multicolumn{2}{c|}{$18.22$}&$62.72$&$10.57$&\multicolumn{2}{c}{$6.22$} \\
    \rowcolor{gray!10}\multirow{-2}{*}{Gemini-1.5-pro}& \mymethod{} &$68.74$\greenup{0.49}&$65.71$\greenup{52.38}&\multicolumn{2}{c|}{$58.45$\greenup{40.23}}&$63.43$\greenup{0.71}&$67.28$\greenup{56.71}&\multicolumn{2}{c}{$62.34$\greenup{56.12}} \\
    \hline

    \Xhline{1.2pt}
  \end{tabular}}
  
  \label{tab:main1}
  
\vspace{-10pt}
\end{table*}

%% file: section/5_exp.tex
We conducted detailed experiments with 21,750 samples across five models in two domains, thoroughly evaluating the performance and privacy effects of both the baseline methods and \mymethod{}.

\subsection{Experimental Setup}

\noindent \textbf{Datasets and Tasks.} Adhering to \citep{multiagent_feng2023knowledge,multiagent_wang2025learning}, we evaluated the performance and privacy of the models in the financial and medical scenarios. Our dataset is divided into three categories: user profiles, multiple-choice questions, and open-ended contextual questions. The detailed generation process of these categories is provided in \Cref{sec:data generation}.

\noindent \textbf{Evaluation Metric.} The structure of a test sample is \(s = \langle \texttt{entry}, \texttt{field}, \texttt{type}, \texttt{stem}, \texttt{answer} \rangle\). We denote the answer obtained by MAS as \(y_{\text{pred}}\) and the pre-defined standard answer as \(y_a\). Due to the difficulty of standardizing reference answers for OEQ across large models, as well as the challenges in controlling evaluation metrics, we primarily use MCQ to assess the utility of MAS \citep{multiagent_bagdasarian2024airgapagent}. The calculation method is as follows:

\vspace{-10pt}
\begin{equation}
    \text{Utility} = \frac{\sum_{|S_{\text{type}}| = \text{MCQ}} \mathbb{I}(y_a, y_{\text{pred}})}{|S_{\text{type}}| = \text{MCQ}},
\end{equation}

\vspace{-7pt}
where \(\mathbb{I}(y_a, y_{\text{pred}})\) is an indicator function that returns 1 if \(y_a = y_{\text{MAS}}\) and 0 otherwise. Privacy evaluation takes a more comprehensive approach, utilizing both MCQ and OEQ. In the case of MCQ, a predefined option, \texttt{Refuse to answer}, is included as the standard answer. For OEQ, agents are guided through prompts containing explicit instructions for their responses.

\vspace{-15pt}
\begin{equation}
\left\{
\begin{aligned}
    &\text{Privacy}_{MCQ} = \frac{\sum_{|S_{\text{type}}| = \text{MCQ}} \mathbb{I}(y_a, y_{\text{pred}})}{|S_{\text{type}}| = \text{MCQ}},\\
    &\text{Privacy}_{OEQ} = \frac{\sum_{|S_{\text{type}}| = \text{OEQ}} \mathbb{EM}(y_a, y_{\text{pred}})}{|S_{\text{type}}| = \text{OEQ}},
\end{aligned}
\right.
\end{equation}
where \(\mathbb{EM}(y_a, y_{\text{pred}})\) is an exact match function that returns 1 if the predicted answer \(y_{\text{pred}}\) exactly matches the reference answer \(y_a\), and 0 otherwise.

\vspace{-2pt}

\begin{equation}
\mathbb{EM} = 
\begin{cases} 
1 & \text{if } S_{\text{pred}} = S_{a} \\
0 & \text{otherwise}
\end{cases}
\end{equation}

\vspace{-4pt}

\subsection{Experiment Results}
We adopt a \texttt{3+n} architecture for evaluation. In the main experiment (\Cref{tab:main1}), we fix \texttt{n} to 1 for evaluation. Additionally, we perform ablation studies by replacing the backbone architectures of the entire MAS and specifically focusing on the backbone of the server-side $C_{\mathcal{A}}$. We also investigate the impact of varying the number of privacy-preserving agents $C_{\mathcal{A}}$ deployed on the server.

\looseness = -1
\noindent \textbf{Performance Analysis.} We observed a slight increase in utility in most scenarios, while the Privacy scores improved significantly across all scenarios. Interestingly, \texttt{GPT-o1} exhibited a significantly higher increase in utility compared to other backbones. We attribute this to the strong comprehension capabilities of \texttt{GPT-o1}, which allows for more precise filtering of user profiles and intermediate data flows. In contrast, models with relatively weaker comprehension capabilities, such as \texttt{Gemini-1.5} and \texttt{GPT-3.5-turbo}, exhibit a utility decline under certain scenarios due to their limited ability to handle tasks effectively. However, even in these cases, the improvement in Privacy remains highly significant.

Additionally, we observed an \textit{entries difference} in Privacy scores. Questions associated with certain entries, such as \texttt{annual income}, which are widely recognized as sensitive privacy information, tend to exhibit higher privacy protection compared to other entries. This effect is particularly prominent in high-performing models like \texttt{Claude} and \texttt{GPT-o1}. In contrast, this distinction is less evident in lower-performing LLMs. For example, the Privacy score of \texttt{GPT-4o} on the Baseline is comparable to that of \texttt{GPT-3.5-turbo}.

\begin{figure}[t]
	\centering
    \begin{center}
		\includegraphics[width=0.99\linewidth]{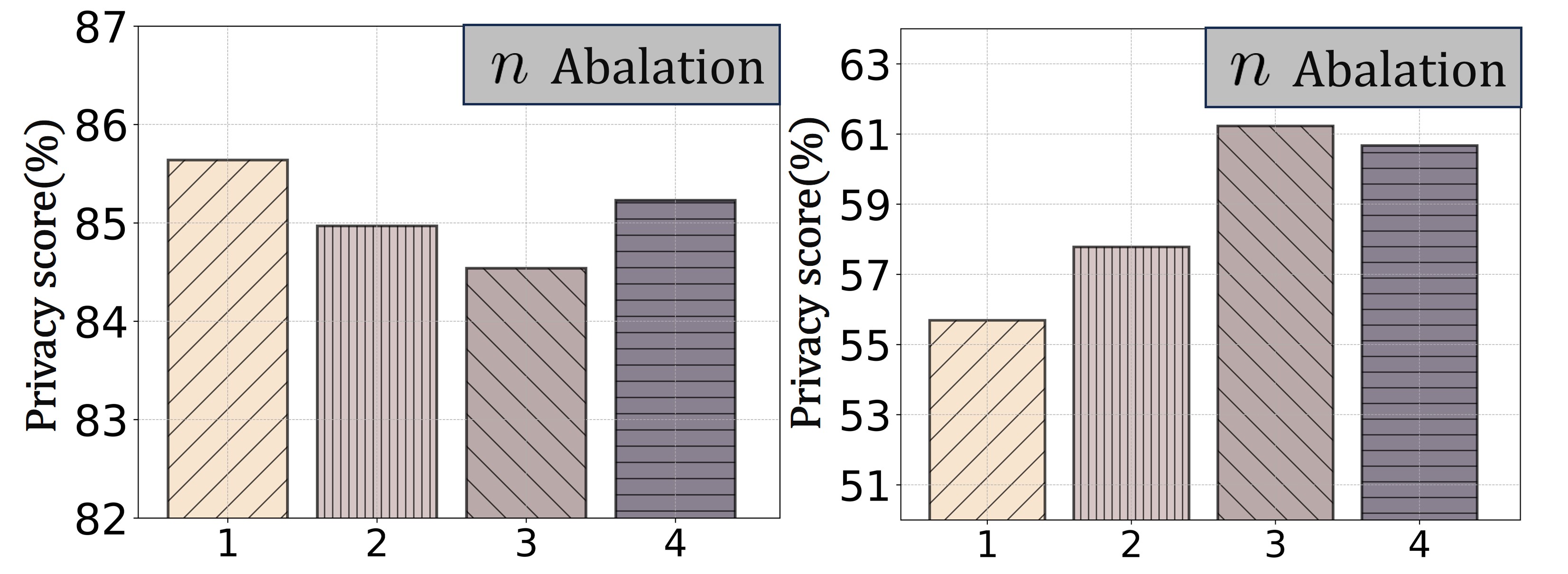}
    \end{center}
    \vspace{-5pt}
    \caption{\small \textbf{Ablation Analysis} of the number of $C_{\mathcal{A}}$. We used \texttt{Claude-3.5} and \texttt{Gemini-1.5} as backbones in our experiments. Please refer to \Cref{sec:abl_ana} for additional analysis.}
    \label{fig:par_aba}
    \vspace{-10pt}
\end{figure}

\vspace{-3pt}
\subsection{Ablation analysis.} 
\noindent \textbf{Different Backbones.} A comparison of columns in \Cref{tab:main1} reveals that the differences in Privacy scores among various backbones in the \textbf{Baseline} are relatively minor. For instance, even the high-performing \texttt{GPT-o1} achieves a Privacy score of only 15.89 in the financial scenario without the application of \mymethod{}, which is merely 3.51\% higher than that of \texttt{GPT-3.5-turbo}. However, when our architecture is applied, the improvement in Privacy scores becomes significantly more pronounced for higher-performing LLMs. For example, \texttt{Claude-3.5} demonstrates a remarkable 71.96\% increase in Privacy scores, whereas \texttt{Gemini-1.5}, being relatively less capable, achieves a more moderate improvement of 44.01\%.

\noindent \textbf{Key Parameters.} We conducted ablation studies on the number of $C_{\mathcal{A}}$ agents deployed on the server to analyze how their workload distribution affects the overall performance of the MAS. The results presented in \Cref{fig:par_aba} show that when lower-performing LLMs are used as the backbone for $C_{\mathcal{A}}$, increasing \texttt{n} slightly improves the Privacy scores. However, this improvement becomes less significant when higher-performing LLMs are used as the backbone. For example, when \texttt{Claude-3.5} is used as the backbone, the Privacy score tends to decrease as \texttt{n} increases. In contrast, with \texttt{Gemini-1.5}, the Privacy score can improve by as much as 6.29\% at its peak.
\label{sec:abl_ana}

\begin{figure}[t]
	\centering
    \begin{center}
		\includegraphics[width=0.99\linewidth]{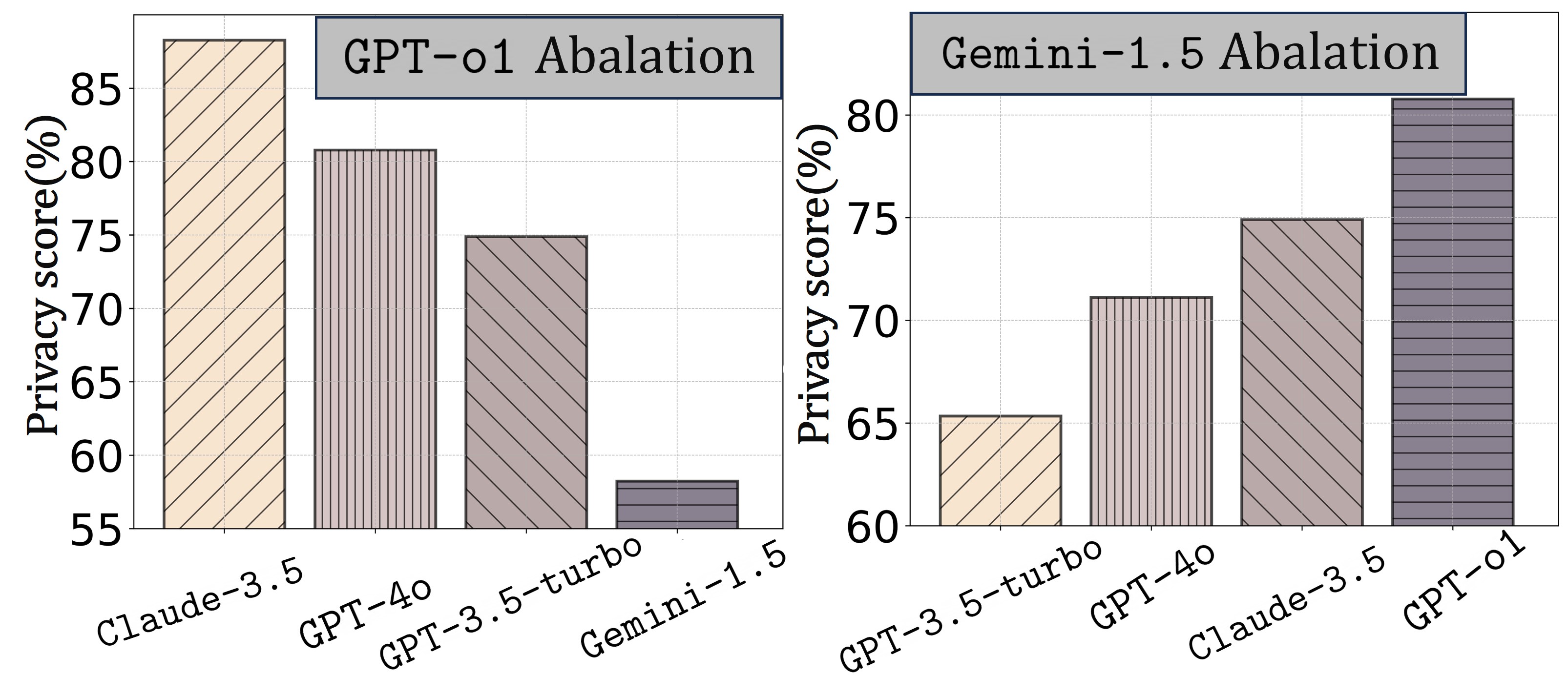}
    \end{center}
    \vspace{-5pt}
    \caption{\small \textbf{Ablation Analysis} of the backbone of $C_{\mathcal{A}}$. We replaced the backbone of $C_{\mathcal{A}}$ with \texttt{GPT-o1} and \texttt{Gemini-1.5} as local agents to study their impact on the privacy score of MAS. Please refer to \Cref{sec:abl_ana} for additional analysis.}
    \label{fig:backbone_abl}
    \vspace{-15pt}
\end{figure}

\noindent \textbf{Backbone of \(C_{\mathcal{A}}\).} WWe conduct ablation studies on the server-side privacy-preserving agent's backbone, focusing on the two models with the best and worst performance in \Cref{tab:main1}: \texttt{GPT-o1} and \texttt{Gemini-1.5}. The results are presented in \Cref{fig:backbone_abl}. Our findings highlight the critical role of the $C_{\mathcal{A}}$ backbone. Even when local agents utilize a high-performing LLM such as \texttt{GPT-o1}, maintaining a high Privacy score becomes challenging if the $C_{\mathcal{A}}$ backbone is suboptimal. For instance, when the backbone of $C_{\mathcal{A}}$ is \texttt{Gemini-1.5}, the Privacy score drops to 58.67\% despite local agents using \texttt{GPT-o1}, representing a 38.95\% decrease from the original score. In contrast, employing a strong LLM as the $C_{\mathcal{A}}$ backbone enables the system to achieve substantial Privacy scores, even when the local agents rely on less capable LLMs. This observation indirectly validates the effectiveness of \mymethod{}.

%% file: section/6_conclusion.tex
\looseness = -1

In this work, we identified emerging privacy protection challenges in LLM-based MAS, particularly within sensitive domains. We introduced the concept of Federated MAS, emphasizing its key distinctions from traditional FL. Addressing critical challenges such as heterogeneous privacy protocols, structural complexities in multi-party conversations, and dynamic conversational network structures, we proposed \mymethod{} as a novel solution. This method minimizes data flow by sharing only task-relevant, agent-specific information and integrates seamlessly into both the RAG and context retrieval stages. Extensive experiments demonstrate \mymethod{}'s potential in real-world applications, providing a robust approach to privacy-preserving multi-agent collaboration. Looking ahead, we highlight the importance of incorporating dynamic privacy-enhancing techniques into MAS, particularly in high-stakes domains where privacy and security are essential.